\documentclass{article} 
\usepackage{iclr2015,times}
\usepackage{hyperref}
\usepackage{url}

\usepackage{subfigure}
\usepackage{epsfig}
\usepackage{graphicx}
\usepackage{amsmath}
\usepackage{amssymb}
\usepackage{bbm}
\usepackage{epstopdf}
\usepackage{enumitem}
\usepackage{calc}
\usepackage{multirow}
\usepackage{color}

\newcommand{\C}{{\cal C}}
\newcommand{\D}{{\cal D}}
\newcommand{\Y}{{\cal Y}}
\newcommand{\X}{{\cal X}}
\newcommand{\R}{{\cal R}}

\def\be {\begin{equation}}
\def\ee {\end{equation}}
\def\beas {\begin{eqnarray*}}
\def\eeas {\end{eqnarray*}}
\def\bea {\begin{eqnarray}}
\def\eea {\end{eqnarray}}
\def\bes {\begin{equation*}}
\def\ees {\end{equation*}}
\def\ba {\begin{align}}
\def\ea {\end{align}}
\def\barr {\begin{array}}
\def\earr {\end{array}}

\newtheorem{lemma-ap}{Lemma}

\newtheorem{claim}{Claim}
\newtheorem{claim-ap}{Claim}

\makeatletter
\usepackage{xspace}
\def\@onedot{\ifx\@let@token.\else.\null\fi\xspace}
\DeclareRobustCommand\onedot{\futurelet\@let@token\@onedot}

\newcommand{\figref}[1]{Fig\onedot~\ref{#1}}
\newcommand{\equref}[1]{Eq\onedot~\eqref{#1}}

\newcommand{\tabref}[1]{Table~\ref{#1}}

\def\eg{\emph{e.g}\onedot} 
\def\ie{\emph{i.e}\onedot} 
 
\def\etc{\emph{etc}\onedot} \def\vs{\emph{vs}\onedot}
\def\wrt{w.r.t\onedot}

\makeatother

\title{Learning  Deep Structured Models}

\author{
Liang-Chieh Chen $^{\ast}$\\
Department of Computer Science, UCLA, \texttt{lcchen@cs.ucla.edu}\\
\And
Alexander~G. Schwing \thanks{The first two authors contributed
  equally to this work.} \\
Department of Computer Science, University of Toronto, \texttt{aschwing@cs.toronto.edu}\\
\And
Alan~L. Yuille\\
Department of Statistics, UCLA, \texttt{yuille@stat.ucla.edu}\\
\And
Raquel Urtasun\\
Department of Computer Science, University of Toronto, \texttt{urtasun@cs.toronto.edu}\\
}

\iclrfinalcopy 

\begin{document}

\maketitle

\begin{abstract}
Many problems in real-world applications involve predicting several random variables which are statistically related.
Markov random fields (MRFs) are a great mathematical tool to encode such relationships. 
The goal of this paper is to combine MRFs with deep learning  algorithms to estimate complex representations while taking into account the dependencies between the output random variables. 
Towards this goal, we propose a training algorithm that  is able to learn structured models jointly with deep features that form the MRF potentials. 
Our approach is efficient as it  blends learning and inference and makes use of GPU acceleration. 
We demonstrate the effectiveness of our algorithm in the tasks of predicting words from noisy images, as well as   multi-class  classification of Flickr photographs. We  show that joint learning of the deep features and the MRF parameters results in significant performance gains.  

\end{abstract}

\section{Introduction}

Deep learning algorithms attempt to model high-level abstractions of the  data  using architectures composed of multiple non-linear transformations.  
 A multiplicity of variants 
have been proposed~\citep{Hinton1984,LeCun1998,HintonScience2006,Bengio2007,Salakhutdinov2012,ZeilerECCV2014} 
and shown to be extremely successful in a wide variety of applications including object detection, speech recognition as well as natural language processing \citep{LeeICML2009,SocherNIPS2012,Jia13caffe,KrizhevskyNIPS2013,EigenICLR2014}. 
Recently,  state-of-the-art results have been achieved  in many computer vision tasks, outperforming competitive methods by a large margin \citep{KrizhevskyNIPS2013,Girshick2014RCNN}. 

Deep neural networks can, however, be even more powerful when combined with graphical models in order to capture the statistical dependencies between the variables of interest.
For example, \citet{Deng14} exploit mutual exclusion, overlapping and subsumption properties of class labels in order to better predict in large scale classification tasks. 
In pose estimation, more accurate predictions can be obtained when encoding the spatial relationships between joint locations \citep{tompson2014joint}.

It is, however,  an open problem how to develop scalable deep learning algorithms that can learn higher-order knowledge taking into account the output variable's dependencies. 
Existing approaches  often rely on a  two-step process \citep{NowozinICCV2011,JiaXu14} where   a non-linear  classifier that employs deep features is trained first, and its output is used to generate  potentials for the structured predictor. 
This piece-wise training is, however,  suboptimal as the deep features are learned while ignoring the dependencies between the variables of interest, \eg,  independently learned segmentation and detection features \citep{hariharan2014simultaneous} might be focusing on predicting the same examples correctly. But when learned jointly they can improve their predictive power by exploiting complementary information to  fix additional mistakes. 


In this paper we extend deep learning  algorithms to  learn complex representations taking into account the dependencies between the output random variables. 
Towards this goal, we propose a learning algorithm that  is able to learn structured models with arbitrary graphs jointly with deep features that form Markov random field (MRF) potentials. 
Our approach is efficient as it  blends learning and inference resulting in a single loop algorithm which makes use of GPU acceleration. 
We demonstrate the effectiveness of our method in the tasks of predicting words from noisy images, and   multi-class  classification  of Flickr photographs. We  show that joint learning of deep features and MRF parameters results in big performance gains.  
\section{Learning Deep Structured Models}


In this section we investigate how to learn `deep features' that take into account the dependencies between the output variables. 
Let $y\in\Y$ be the set of random variables $y = (y_1, \ldots, y_N)$ that we are interested in predicting. 
In this work we assume the space of valid configurations to be a product space, \ie, $\Y = \prod_{i=1}^N \Y_i$, and the domain of each individual variable $y_i$ to be  discrete, \ie,   $\Y_i = \{1, \ldots, |\Y_i|\}$.
Given  input data $x\in\X$ and  parameters $w\in\mathbb{R}^A$ of the  function $F(x,y;w) : \X\times\Y\times\mathbb{R}^A \rightarrow \mathbb{R}$, inference amounts to finding the highest scoring  configuration 
$
  y^\ast = \arg \max_{y} F(x,y;w).
$
Note that if $F$ is a deep network and there are  no connections between the output variables to be predicted, inference corresponds to a forward pass to evaluate the function, followed by  independently finding the largest response for each variable. This can be interpreted as inference in  a graphical model with only unary potentials. 
However, for arbitrary graphical models it is NP-hard to find the maximizing configuration $y^\ast$ since the inference program generally requires a search over a space of size $\prod_{i=1}^N |\Y_i|$. 
Note also that log-linear models are  a special case of this program, with  $F(x,y;w) = w^\top \phi(x,y)$ and $\phi(x,y)$ denoting a feature vector computed using the input-output pair $(x,y)$. 

In this work, we  consider the  general setting where $F(x,y;w)$ is an arbitrary scalar-valued function of $w$ and $(x,y)$. 
In our experiments $F$ is a function composition of non-linear base mappings like convolutions, rectifications, pooling \etc. 
We let the probability of an arbitrary configuration $\hat y$ be given by the annealed soft-max
$
  p_{(x,y)}(\hat{y} | w,\epsilon) = \frac{1}{Z_\epsilon(x,w)} \exp(F(x,\hat{y};w))^{1/\epsilon}.
$
Hereby  $Z_\epsilon(x,w)$ refers to the partition function, normalizing the distribution $p_{(x,y)}$ to lie within the probability simplex $\Delta$ via $Z(x,w) = \sum_{\hat y \in \Y} \exp(F(x,\hat y;w))^{1/\epsilon}$. The annealing/temperature parameter $\epsilon \geq 0$ is used to adjust the uniformity of the distribution.  We consider general graphical models where the computation of $Z_\epsilon(x,w)$ is  \#P-hard. 

\subsection{Learning via gradient descent}

During learning, given a training set $\D$ of input-output pairs $(x,y)\in\D$, we are interested in  finding the parameters $w$ of the model. 
We do so by  maximizing the data likelihood, \ie, minimizing the negative log-likelihood $-\ln \prod_{(x,y)\in\D} p_{(x,y)}(y|w,\epsilon)$ which yields
\be
  \min_{w} \sum_{(x,y) \in \D}\left(\epsilon\ln Z_\epsilon(x,w) - F(x,y;w) \right).
	\label{eq:Learning}
\ee
Note that this is equivalent to maximizing the cross-entropy between a target distribution $p_{(x,y),\operatorname{tg}}(\hat y) = \delta(\hat y = y)$ placing all its mass on the groundtruth label, and the model distribution $p_{(x,y)}(\hat y | w, \epsilon)$. Hence \equref{eq:Learning} is equivalently obtained by $\max_{w} \sum_{(x,y),\hat y \in \Y} p_{(x,y),\operatorname{tg}}(\hat y)\ln p_{(x,y)}(\hat y|w, \epsilon)$. It is easily possible to incorporate more general target distributions into \equref{eq:Learning}.
Note also that $\epsilon=0$ recovers the structured hinge loss objective. 

Minimizing  \equref{eq:Learning} \wrt $w$ requires computation of the gradient $\frac{\partial}{\partial w} \sum_{(x,y)\in\D}-\ln p_{(x,y)}(y|w,\epsilon)$, which is given by  a transformed difference between the distributions of the model  $p_{(x,y)}(\hat y|w,\epsilon)$ and  the target  $p_{(x,y),\operatorname{tg}}(\hat y)$: 
\begin{equation}
\sum_{(x,y) \in \D} \sum_{\hat y\in\Y} \frac{\partial}{\partial w} F(x,\hat y;w)\left(p_{(x,y)}(\hat y|w,\epsilon) - p_{(x,y),\operatorname{tg}}(\hat y)\right).
\label{eq:gradient}
\end{equation}
A gradient descent algorithm for minimizing  \equref{eq:Learning} will iterate between the following steps: (i) For a given  $w$ evaluate the function $F$, (ii) compute the model distribution $p_{(x,y)}(\hat y| w, \epsilon)$, (iii) propagate the difference between the model and target distribution using a backward pass (resembling the chain rule for composite functions) and  (iv) update the parameters $w$. This is summarized in \figref{fig:NonLinStructPredAlgoSummary}.

\subsection{Approximate Learning}

\begin{figure}[t]
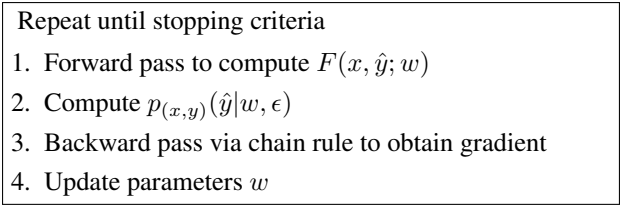

\centering
\fbox{
\begin{minipage}[c]{7.8cm}

Repeat until stopping criteria

\begin{enumerate}[labelindent=0pt,labelwidth=\widthof{\ref{last-item}},label=\arabic*.,itemindent=1em,leftmargin=0pt]
\item Forward pass to compute $F(x,\hat y; w)$
\item Compute $p_{(x,y)}(\hat y | w, \epsilon)$ 
\item Backward pass via chain rule to obtain gradient
\item Update parameters $w$
\end{enumerate}

\end{minipage}
}
\caption{Gradient descent algorithm for learning deep structured models.}
\label{fig:NonLinStructPredAlgoSummary}
\end{figure}


Note that for general graphical models the exact computation of $p_{(x,y)}(\hat y| w, \epsilon)$ is not possible.
As a consequence it is intractable to compute the exact gradient of the cost-function given in \equref{eq:gradient} and  one has to resort to approximate solutions. 

Inspired by approximations used for log-linear models, 
we make use of the following identity~\citep{Wainwright2008,Koller2009}:
\begin{align}
  \label{eqn:log_partition}
  \epsilon\ln Z_{\epsilon}(x,w) = \max_{p_{(x,y)}(\hat{y}) \in \Delta} \mathbb{E} [F(x,\hat{y};w) ] + \epsilon H(p_{(x,y)}),
\end{align}
where $\mathbb{E}$ denotes an expectation over $p_{(x,y)}(\hat y)$ and $H$ refers to the entropy. 

For most applications,  $F(x,y;w)$  decomposes into a sum of functions, each depending on a local subset of variables $y_r$, \ie,
$
  F(x,y;w) = \sum_{r\in\R} f_r(x,y_r;w).
$
Hereby $r$ is a restriction of the variable tuple $y = (y_1, \ldots, y_N)$ to the subset $r\subseteq\{1, \ldots, N\}$, \ie, $y_r = (y_i)_{i\in r}$. All subsets $r$ required to compute the model function $F$ are summarized in the set $\R$.

Plugging this decomposition into  \equref{eqn:log_partition}, we equivalently get the log-partition function $\epsilon \ln Z_\epsilon(x,w)$ via 
$
\max_{p_{(x,y)}(\hat y) \in \Delta}\sum_{r,\hat y_r} p_{(x,y),r}(\hat y_r)f_r(x,\hat y_r;w) + \epsilon H(p_{(x,y)}),
$
where we use  marginals $p_{(x,y),r}(\hat y_r) = \sum_{y\setminus y_r} p_{(x,y)}(y)$.

Despite the assumed locality of the scoring function, the learning task remains computationally challenging since the entropy $H(p_{(x,y)})$ can only be computed exactly for a very small set of models, \eg, models for which the joint distribution $p_{(x,y)}(y)$ is equivalently described by low tree-width models. 
In addition, the marginalization constraints are exponential in size. 

To deal with both issues a common solution in log-linear models is to approximate the true marginals $p_{(x,y),r}$ with local beliefs $b_{(x,y),r}$ that are not required to fulfill marginalization constraints globally, but only locally \citep{Wainwright2008}.  That means marginals $b_{(x,y),r}$ are not required to arise from a common joint distribution $p_{(x,y)}$. In addition, we approximate the entropy via the fractional entropy~\citep{Wiegerinck2003}, \ie, $H(p_{(x,y)}) \approx \sum_r c_r H(b_{(x,y),r})$. Counting numbers $c_r$ are employed to weight the marginal entropies. Putting all this together,  we obtain the following approximation for $\epsilon \ln Z_\epsilon(x,w)$: 
\be
\max_{b_{(x,y)}\in\mathcal{C}_{(x,y)}} \sum_{r,\hat y_r} b_{(x,y),r}(\hat y_r)f_r(x,\hat y_r;w) + \sum_r \epsilon c_r H(b_{(x,y),r}).
\label{eq:approxInf}
\ee
Hereby beliefs are constrained to the local polytope
$$
\mathcal{C}_{(x,y)} = \left\{\barr{ll}\forall r & b_{(x,y),r}\in\Delta\\\forall r,\hat y_r,p\in P(r) & \sum\limits_{\hat y_p\setminus \hat y_r} \!\!\!b_{(x,y),p}(\hat y_p) = b_{(x,y),r}(\hat y_r),\earr\right.
$$
with $P(r)$ the set of parents of region $r$, \ie, $P(r) \subseteq \{p\in\R : r \subset p\}$, which subsumes those regions for which we want the marginalization constraint to hold. Conversely, we define the set of children as $C(r) = \{c\in\R : r\in P(c)\}$. 


We can thus rewrite the learning problem by plugging the approximations derived in \equref{eq:approxInf} into \equref{eq:Learning}. This gives rise to the new approximated learning program 
\be
\min_w \!\!\sum_{(x,y)\in\D}\!\! \left(\!\!\max_{b_{(x,y)}\in\mathcal{C}_{(x,y)}}\!\! \left\{\sum_{r,\hat y_r} b_{(x,y),r}(\hat y_r)f_r(x,\hat y_r;w) + \sum_r\epsilon c_r H(b_{(x,y),r})\right\} - F(x,y;w)\right).
\label{eq:structPredGeneral}
\ee

To iteratively update the parameters for the non-smooth approximated cost function given in \equref{eq:structPredGeneral}
we require a sub-gradient \wrt $w$, which  in turn  requires to solve the maximization \wrt the beliefs $b$  exactly. This is a non-trivial task in itself as inference in general graphical models is NP-hard. Iterative message passing algorithms \citep{Pearl1988,Yedidia2005,Wainwright2005b,Weiss2007,Meltzer2009} are typically employed. 
Importantly, note that  combining the procedure outlined in \figref{fig:NonLinStructPredAlgoSummary} with iterative message passing to approximate $p_{(x,y)}(\hat y | w, \epsilon)$ results in a double-loop algorithm which would be slow for many graphical models of  interest. 



\begin{figure*}[t]
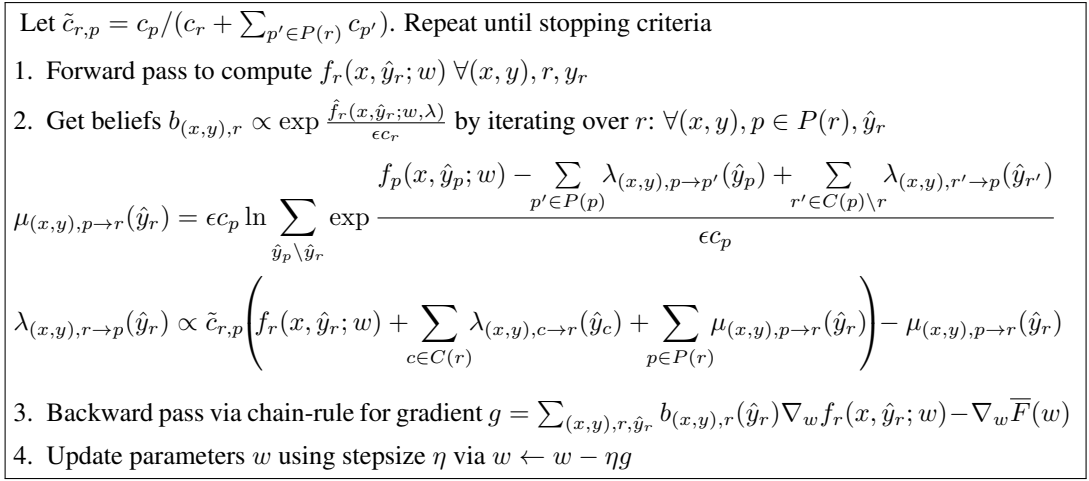

\fbox{
\begin{minipage}[c]{\linewidth}

Let $\tilde{c}_{r,p} = c_p/(c_r + \sum_{p^\prime\in P(r)} c_{p^\prime})$. Repeat until stopping criteria

\begin{enumerate}[labelindent=0pt,labelwidth=\widthof{\ref{last-item}},label=\arabic*.,itemindent=1em,leftmargin=0pt]
\item Forward pass to compute $f_r(x,\hat y_r; w)$ $\forall (x,y),r,y_r$
\item Get beliefs $b_{(x,y),r} \propto \exp\frac{\hat f_r(x,\hat y_r;w,\lambda)}{\epsilon c_r}$ by iterating over $r$: $\forall (x,y),p\in P(r), \hat y_r$
\bea
&&\hspace{-0.9cm} \mu_{(x,y),p\rightarrow r}(\hat y_r) = \epsilon c_p \ln \sum_{\hat y_p\setminus \hat y_r} \exp \frac{f_p(x,\hat y_p;w) - \!\!\sum\limits_{p^\prime\in P(p)}\!\!\lambda_{(x,y),p\rightarrow p^\prime}(\hat y_p) + \!\!\sum\limits_{r^\prime\in C(p)\setminus r}\!\! \lambda_{(x,y),r^\prime\rightarrow p}(\hat y_{r^\prime})}{\epsilon c_p}\nonumber
\\
&&\hspace{-0.9cm} \lambda_{(x,y),r\rightarrow p}(\hat y_r) \propto \tilde{c}_{r,p} \!\!\left(\!\!f_r(x,\hat y_r;w) + \!\!\sum_{c\in C(r)}\!\! \lambda_{(x,y),c\rightarrow r}(\hat y_c) + \!\!\sum_{p\in P(r)}\!\! \mu_{(x,y),p\rightarrow r}(\hat y_r)\!\!\right)\!\! - \mu_{(x,y),p\rightarrow r}(\hat y_r)\nonumber 
\eea
\item Backward pass via chain-rule for gradient
$
g = \sum_{(x,y),r,\hat y_r} b_{(x,y),r}(\hat y_r)\nabla_w f_r(x,\hat y_r;w) - \nabla_w \overline{F}(w)
$
\item Update parameters $w$ using stepsize $\eta$ via\label{last-item}
$
w \leftarrow w - \eta g
$

\end{enumerate}

\end{minipage}
}
\vspace{-0.0cm}
\caption{Efficient learning algorithm that blends learning and inference. 
}
\label{fig:NonLinStructPredAlgo}
\end{figure*}

\subsection{Efficient Approximate Learning by Blending Learning and Inference}
\label{sec:blended}

In this section we  propose a more efficient  algorithm that is based on the principle of blending learning (\ie, parameter updates) and inference. 
Thus  we are interested in only performing a single message passing iteration before updating the parameters $w$. Note that simply reducing the number of iterations is generally not an option as the obtained beliefs $b_{(x,y),r}$ are by no means accurate. However, assuming all counting numbers $c_r$ to be positive, we can derive an algorithm that is able to interleave minimization \wrt $w$ and maximization of the beliefs $b$. Such a procedure is more efficient as we are able to update the parameters $w$ much more frequently.

To interleave both programs  we convert maximization of the beliefs into a minimization by employing the dual program as detailed in the following claim. 
This is possible since the maximization problem is concave in $b_{(x,y)}$ if $\forall r$, $\epsilon c_r \geq 0$. 
\begin{claim}
Assume $\epsilon c_r \geq 0$ $\forall r$ and let $\overline{F}(w) = \sum_{(x,y)\in\D} F(x,y;w)$ denote the sum of empirical function observations. 
Let  $\lambda_{(x,y),r\rightarrow p}(\hat y_r)$ be the Lagrange multipliers for each marginalization constraint $\sum_{\hat y_p\setminus\hat y_r} b_{(x,y),p}(\hat y_p) = b_{(x,y),r}(\hat y_r)$ within the polytope $\C_{(x,y)}$. Then the approximated general structured prediction task shown in \equref{eq:structPredGeneral} is equivalent to 
\be
\min_{w,\lambda} \sum_{(x,y),r} \epsilon c_r \ln\sum_{\hat y_r}\exp\frac{\hat f_r(x,\hat y_r;w,\lambda)}{\epsilon c_r} - \overline{F}(w),
\label{eq:structPredGeneralDual}
\ee
where we employed the re-parameterization score $\hat f_r(x,\hat y_r;w,\lambda) = f_r(x,\hat y_r;w) + \!\!\sum\limits_{c\in C(r)}\!\!\lambda_{(x,y),c\rightarrow r}(\hat y_c) - \!\!\sum\limits_{p\in P(r)}\!\! \lambda_{(x,y),r\rightarrow p}(\hat y_r)$.
\end{claim}
{\bf Proof:} To obtain the dual of the maximization \wrt $b_{(x,y)}$ we utilize its Lagrangian 
$
L_{(x,y)} 
\!\!=\!\! \sum_{r,\hat y_r}\! b_{(x,y),r}(\hat y_r)\hat f_r(x,\hat y_r;w,\lambda) + \!\sum_{r}\!\epsilon c_r H(b_{(x,y),r}).
$
Maximization of the Lagrangian \wrt the primal variables $b$ is possible by employing the relationship stated in \equref{eqn:log_partition} locally $\forall r$. We then obtain the dual function being the first term in \equref{eq:structPredGeneralDual}.
For strict convexity, \ie, $\epsilon c_r > 0$, we reconstruct the beliefs to be proportional to the exponentiated, loss-augmented re-parameterization score 
$
b_{(x,y),r} \propto \exp \frac{\hat f_r(x,\hat y_r;w,\lambda)}{\epsilon c_r}.
$
For $\epsilon c_r = 0$ the beliefs correspond to a uniform distribution over the set of maximizers of the loss-augmented re-parameterization score $\hat f_r(x,\hat y_r;w,\lambda)$. 
\hfill$\blacksquare$ 

It is important to note that by applying duality we managed to convert the $\min$-$\max$ task in \equref{eq:structPredGeneral} into a single minimization as shown in \equref{eq:structPredGeneralDual}.  Performing block coordinate descent updates to minimize \equref{eq:structPredGeneralDual}, we are therefore able to interleave both, updating the weights (\ie, learning) and the messages (\ie, inference). This results in a more efficient algorithm, as inference does not have to be run until convergence. Even a single  update of the messages suffices. 
We note that this is possible only if $\epsilon c_r \geq 0$ $\forall r$. 
Strictly speaking, we require concavity only within the set of feasible beliefs $\C_{(x,y)}$. 
However, for simplicity we neglect this extension in the following. 

 \figref{fig:NonLinStructPredAlgo}  summarizes 
our  efficient deep structured prediction algorithm which iterates between the following steps. 
Given parameters $w$   we perform a standard forward pass to compute $f_r(x,\hat y_r; w)$ for all regions. We then iterate through all regions $r$ and use block-coordinate descent to find the globally optimal value of \equref{eq:structPredGeneralDual} \wrt $\lambda_{(x,y),r \rightarrow p}(\hat y_r)$ $\forall (x,y), \hat y_r, p\in P(r)$. This can be done   in  closed form and therefore is computed very efficiently. We refer the reader to \citet{Schwing2013} for a derivation in the log-linear setting. We then compute the gradient using a standard backward pass before we update the parameters by performing a step of size $\eta$ along the negative gradient. 

\begin{figure*}[t]
\vspace{-0.0cm}
  \centering
  \begin{tabular}{c  c  c}
    \addtolength{\tabcolsep}{-6.5pt}
  {
    \begin{tabular}{c c c c c}
    \includegraphics[height=0.06\linewidth]{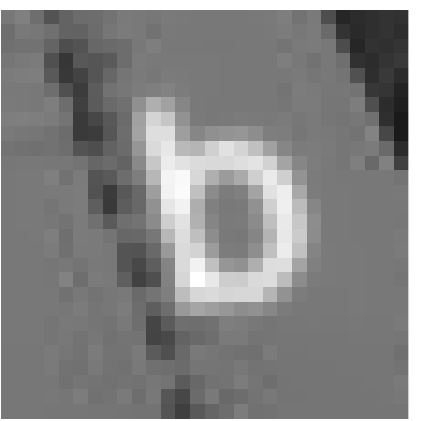} &
    \includegraphics[height=0.06\linewidth]{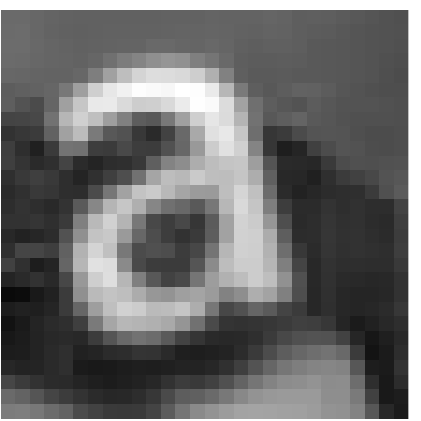} &
    \includegraphics[height=0.06\linewidth]{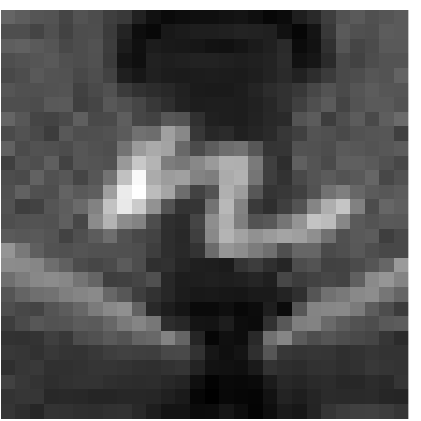} &
    \includegraphics[height=0.06\linewidth]{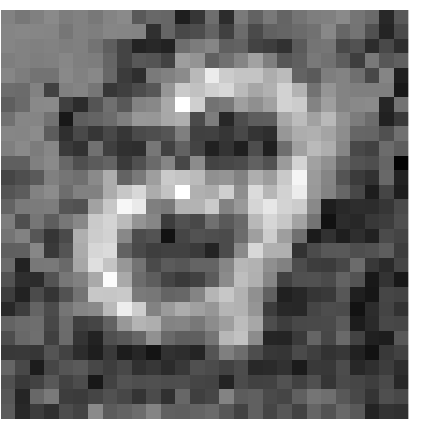} &
    \includegraphics[height=0.06\linewidth]{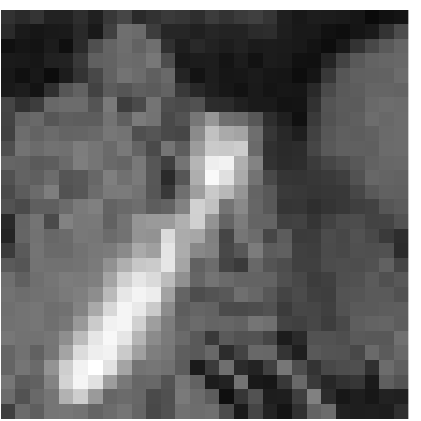} \\
  \end{tabular} 
  } &
  \addtolength{\tabcolsep}{-6.5pt}
  {
  \begin{tabular}{c c c c c}
    \includegraphics[height=0.06\linewidth]{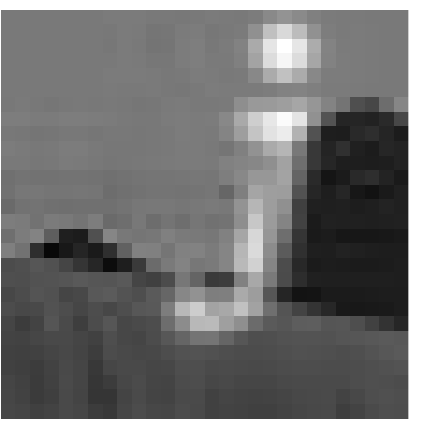} &
    \includegraphics[height=0.06\linewidth]{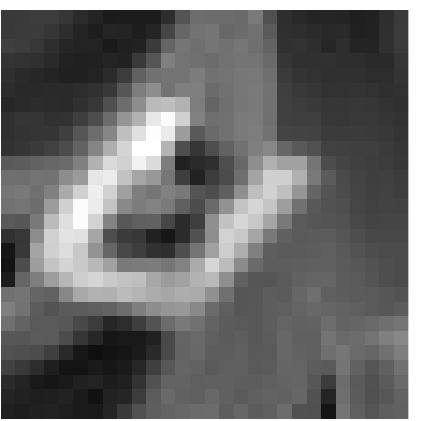} &
    \includegraphics[height=0.06\linewidth]{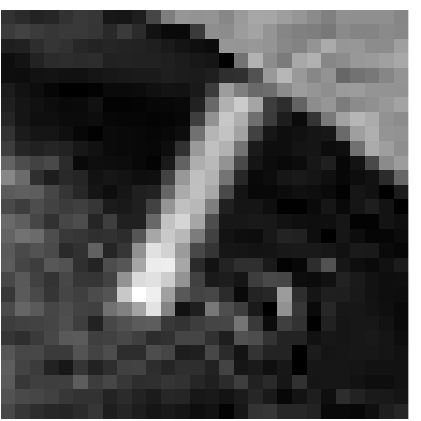} &
    \includegraphics[height=0.06\linewidth]{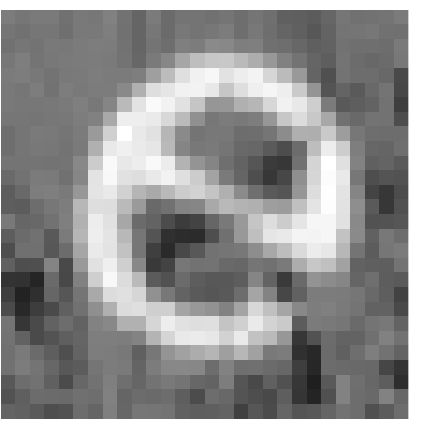} &
    \includegraphics[height=0.06\linewidth]{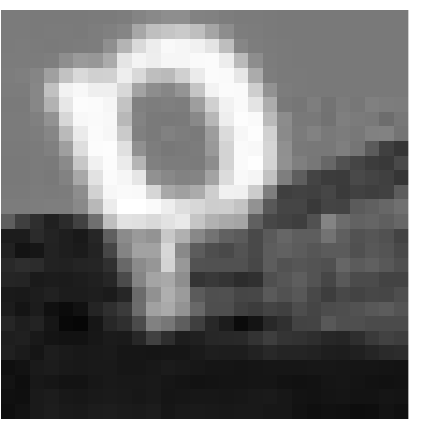} \\ 
  \end{tabular}
  } &
  \addtolength{\tabcolsep}{-6.5pt}
  {
  \begin{tabular}{c c c c c}
    \includegraphics[height=0.06\linewidth]{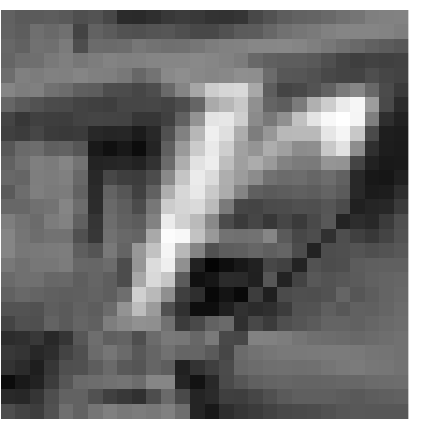} &
    \includegraphics[height=0.06\linewidth]{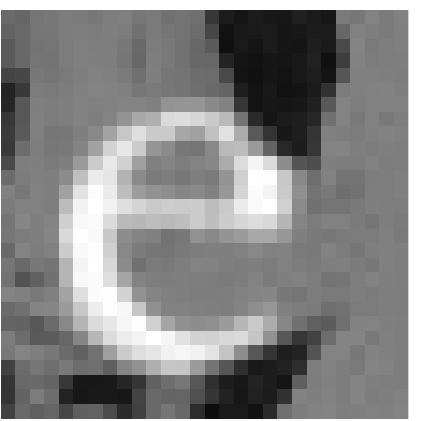} &
    \includegraphics[height=0.06\linewidth]{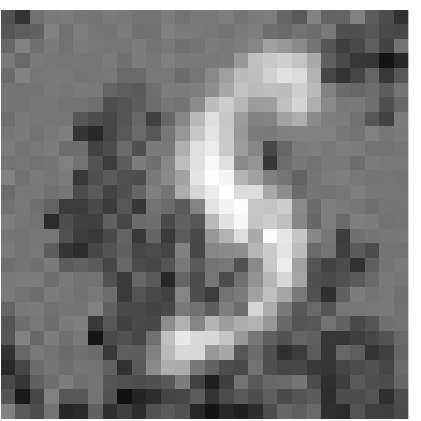} &
    \includegraphics[height=0.06\linewidth]{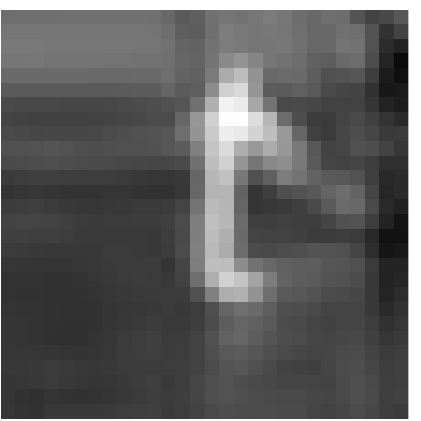} &
    \includegraphics[height=0.06\linewidth]{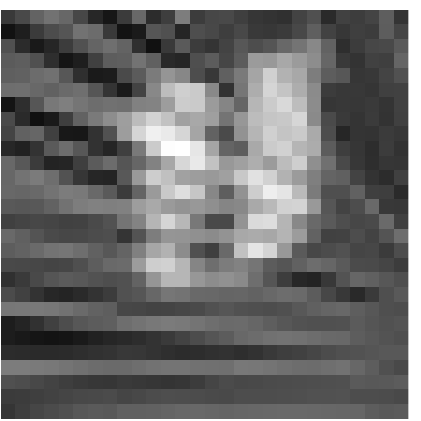} \\ 
  \end{tabular}
  }  \\
  banal & julep & resty\\

  \addtolength{\tabcolsep}{-6.5pt}
  {
    \begin{tabular}{c c c c c}
    \includegraphics[height=0.06\linewidth]{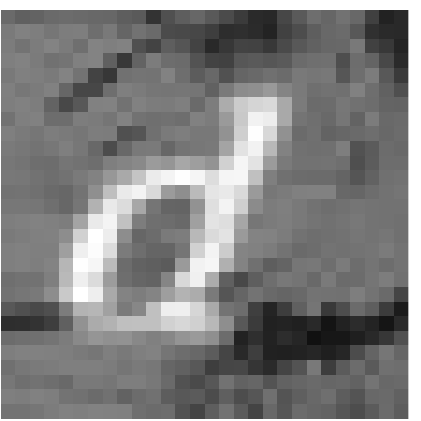} &
    \includegraphics[height=0.06\linewidth]{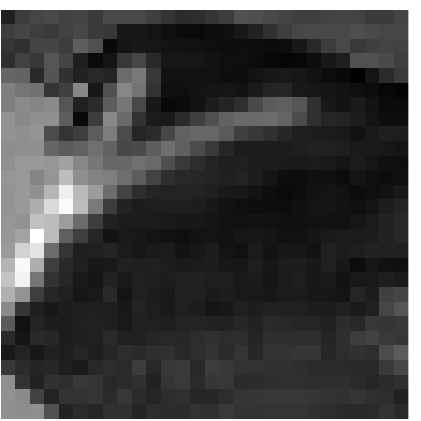} &
    \includegraphics[height=0.06\linewidth]{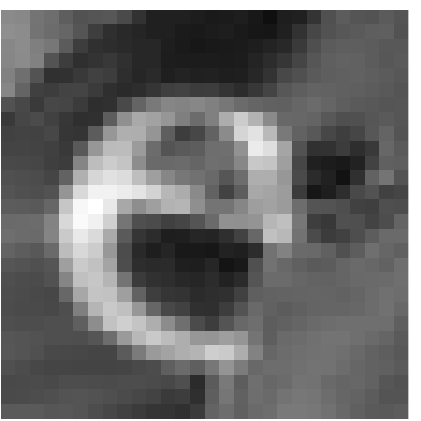} &
    \includegraphics[height=0.06\linewidth]{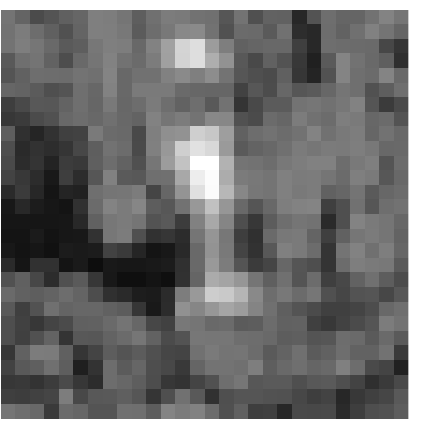} &
    \includegraphics[height=0.06\linewidth]{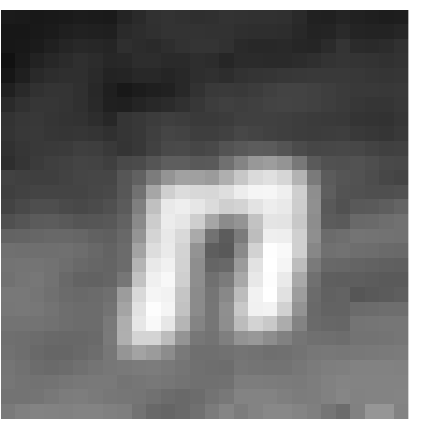} \\
  \end{tabular} 
  } &
  \addtolength{\tabcolsep}{-6.5pt}
  {
  \begin{tabular}{c c c c c}
    \includegraphics[height=0.06\linewidth]{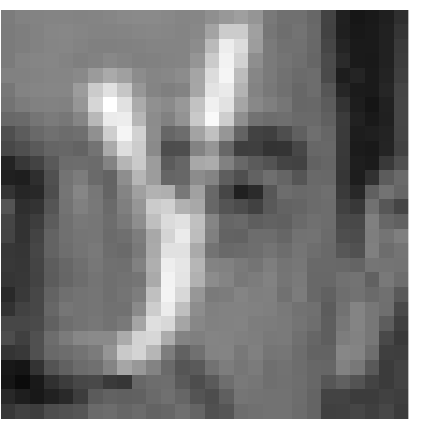} &
    \includegraphics[height=0.06\linewidth]{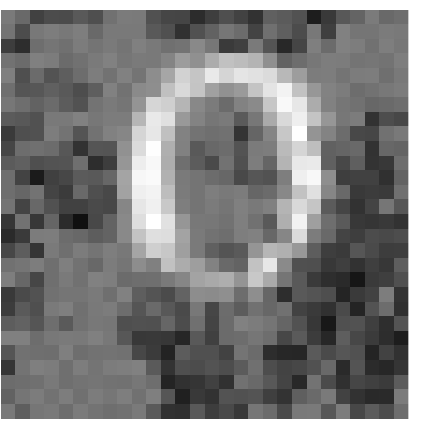} &
    \includegraphics[height=0.06\linewidth]{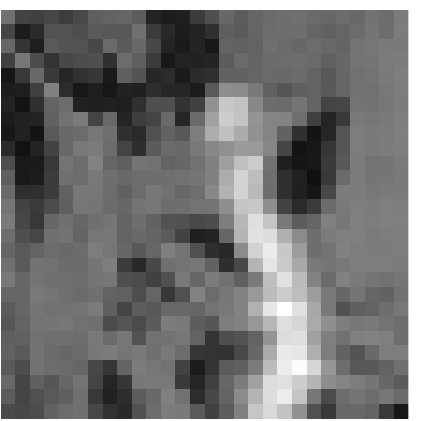} &
    \includegraphics[height=0.06\linewidth]{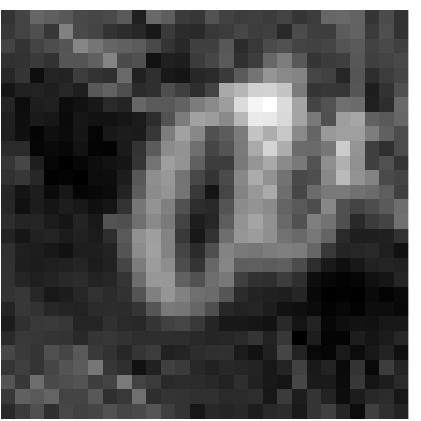} &
    \includegraphics[height=0.06\linewidth]{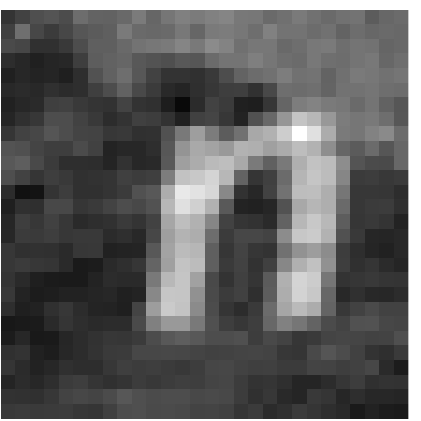} \\ 
  \end{tabular}
  } &
  \addtolength{\tabcolsep}{-6.5pt}
  {
  \begin{tabular}{c c c c c}
    \includegraphics[height=0.06\linewidth]{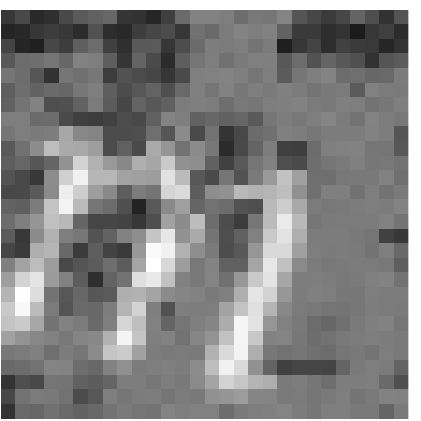} &
    \includegraphics[height=0.06\linewidth]{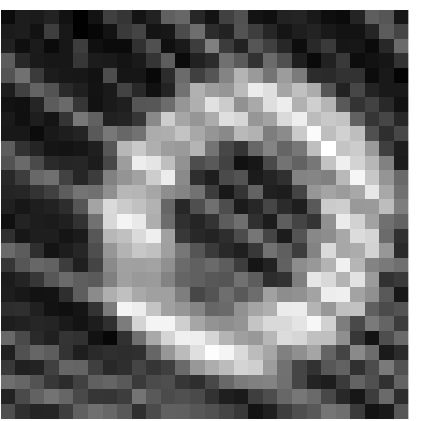} &
    \includegraphics[height=0.06\linewidth]{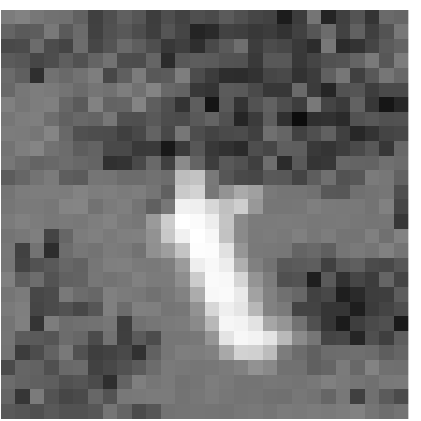} &
    \includegraphics[height=0.06\linewidth]{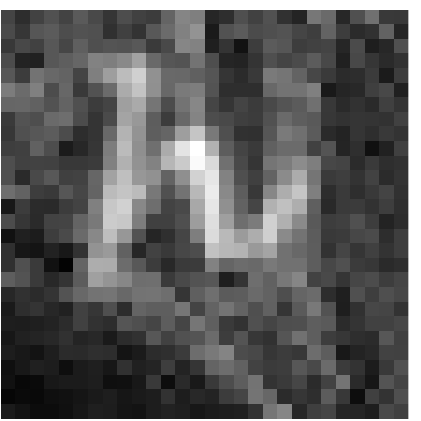} &
    \includegraphics[height=0.06\linewidth]{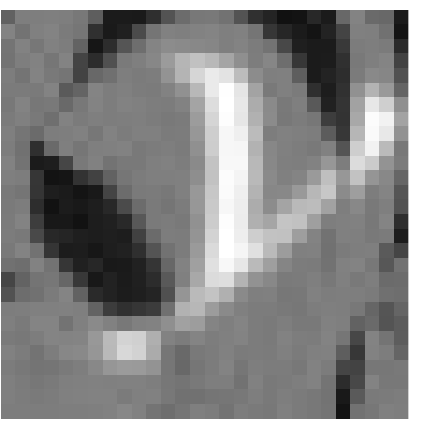} \\ 
  \end{tabular}
  }  \\
  drein & yojan & mothy\\

  \end{tabular}
  \vspace{-0.0cm}
  \caption{Samples from the Word50 dataset. High degree of rotation, scaling and translation. 
  }
	\label{fig:WordDataSet}
\end{figure*}

\subsection{Implementation Details}
We implemented the general algorithm presented in \figref{fig:NonLinStructPredAlgo} in C++ as a library for Linux, Windows and OS X platforms. It supports usage of the GPU for the forward and backward pass using both, standard linear algebra packages and manually tuned GPU-kernels. In addition to standard gradient descent, we allow specification of both mini-batches, moments and different regularizers like $2$-norm and $\infty$-norm. Between iterations the step-size can be reduced based on either the 
negative log-likelihood or validation set performance. Contrasting available deep learning packages, our function $F$ is specified using a general computation tree. Hence we support an arbitrarily nested function structure composed of data, parameters and function prototypes (convolution, affine function aka fully connected, dropout, local response normalization, pooling, rectified linear, sigmoid and softmax units).
The aforementioned library is accompanied by a program performing learning, inference and gradient checks. To accommodate for large datasets it reads data from HDF5 storage while a second thread simultaneously performs the computation. Google protocol buffers are employed to effectively specify the function $F$ without the need to modify any source code.
We will release this library upon publication. We believe that it will be useful for many researchers.

\begin{table*}[t]
\centering
  \scalebox{0.85}{
	\setlength\tabcolsep{1pt}
   \hspace{-1.3mm}\begin{tabular}{|c | c | c | c | c | c | c | c |}
    \hline
    Graph & MLP & Method & $H_1 = 128$ & $H_1 = 256$ & $H_1 = 512$ & $H_1 = 768$ & $H_1 = 1024$ \\
    \hline\hline
    \multirow{4}{*}{1st order Markov} & \multirow{4}{*}{One Layer} &Unary only  & 8.60 / 61.32  & 10.80 / 64.41 & 12.50 / 65.69 & 12.95 / 66.66 & 13.40 / 67.02 \\
    \cline{3-8}
    & &JointTrain                & 16.80 / 65.28  & 25.20 / 70.75 & \textbf{31.80} / 74.90 & 33.05 / 76.42 & 34.30 / 77.02 \\
    \cline{3-8}
    & &PwTrain                 & 12.70 / 64.35  & 18.00 / 68.27 & 22.80 / 71.29 & 23.25 / 72.62 & 26.30 / 73.96 \\   
    \cline{3-8}
    & &PreTrainJoint              & \textbf{20.65} / \textbf{67.42} & \textbf{25.70} / \textbf{71.65} & 31.70 / \textbf{75.56} & \textbf{34.50} / \textbf{77.14} & \textbf{35.85} / \textbf{78.05} \\   
    \hline
    \hline
    \multirow{3}{*}{2nd order Markov} & \multirow{3}{*}{One Layer} & JointTrain    & 25.50 / 67.13 &  34.60 / 73.19 & 45.55 / 79.60 & \textbf{51.55} / \textbf{82.37} & \textbf{54.05} / \textbf{83.57} \\   
    \cline{3-8}
    & &PwTrain           & 10.05 / 58.90  & 14.10 / 63.44 & 18.10 / 67.31 &  20.40 / 70.14 & 22.20 / 71.25 \\  
    \cline{3-8}
    & &PreTrainJoint        & \textbf{28.15} / \textbf{69.07} & \textbf{36.85} / \textbf{75.21}  & \textbf{45.75} / \textbf{80.09}  & 50.10 / 82.30 & 52.25 / 83.39 \\  
    \hline
    \hline
    & & $H_1 = 512$ & $H_2=32$ & $H_2 = 64$ & $H_2 = 128$ & $H_2 = 256$ & $H_2 = 512$ \\
    \hline
    \multirow{4}{*}{1st order Markov} & \multirow{4}{*}{Two Layer} & Unary only & 15.25 / 69.04  &  18.15 / 70.66 & 19.00 / 71.43 & 19.20 / 72.06 & 20.40 / 72.51  \\  
    \cline{3-8}
    & &JointTrain   & 35.95 / 76.92 & 43.80 / 81.64 & 44.75 / 82.22 & 46.00 / 82.96 & \textbf{47.70} / 83.64 \\
    \cline{3-8}
    & &PwTrain    & 34.85 / 79.11 & 38.95 / 80.93 & 42.75 / 82.38 & 45.10 / 83.67 & 45.75 / 83.88 \\
    \cline{3-8}
    & &PreTrainJoint & \textbf{42.25} / \textbf{81.10} & \textbf{44.85} / \textbf{82.96} & \textbf{46.85} / \textbf{83.50} & \textbf{47.95} / \textbf{84.21} & 47.05 / \textbf{84.08} \\ 
    \hline
    \hline
    \multirow{3}{*}{2nd order Markov} & \multirow{3}{*}{Two Layer} & JointTrain & 54.65 / 83.98  & 61.80 / 87.30 & 66.15 / 89.09 & 64.85 / 88.93 & 68.00 / 89.96 \\
    \cline{3-8}
    & &PwTrain    & 39.95 / 81.14 & 48.25 / 84.45 & 52.65 / 86.24  & 57.10 / 87.61 & 62.90 / 89.49 \\ 
    \cline{3-8}
    & &PreTrainJoint  & \textbf{62.60} / \textbf{88.03} & \textbf{65.80} / \textbf{89.32} & \textbf{68.75} / \textbf{90.47} & \textbf{68.60} / \textbf{90.42} & \textbf{69.35} / \textbf{90.75}  \\
    \hline
   \end{tabular}
  }
    \vspace{-0.0cm}  
\caption{Word / Character accuracy.  Performance improves as (1) joint-training is employed, (2) the model is more structured, and (3) deeper unary classifiers are utilized. The number of hidden units for the first  and second layer are denoted as $H_1$ and $H_2$ respectively.}
\label{tab:ExpWordsDist1}

\end{table*}

\section{Experimental Evaluation}
We demonstrate the performance of our model on two tasks: word recognition and image classification. 
We investigate four strategies to learn the model parameters.  `{\bf Unary only}' denotes training only unary classifiers while ignoring the structure of the graphical model, \ie, pairwise weights are equal to $0$.   `{\bf JointTrain}'  initializes all weights at random 
and trains them jointly.  `{\bf PwTrain}' uses piecewise training by first training the unary  potentials and then  keeping them fixed when learning the pairwise potentials.  `{\bf PreTrainJoint}'  pre-trains the unaries but jointly optimizes pairwise weights as well as unary weights in a second step.

\begin{figure*}[t]
\vspace{-0.0cm}
  \centering
  \begin{tabular}{c c c c}  
    \includegraphics[height=0.17\linewidth]{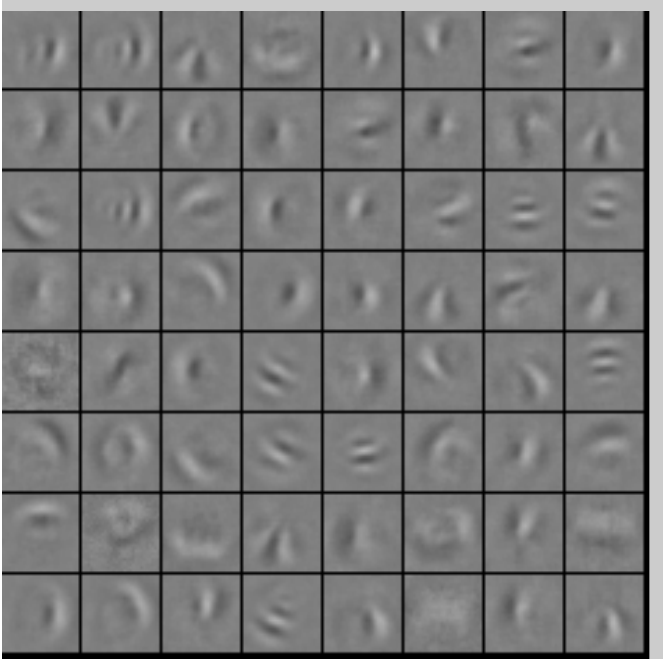} &
    \includegraphics[height=0.17\linewidth]{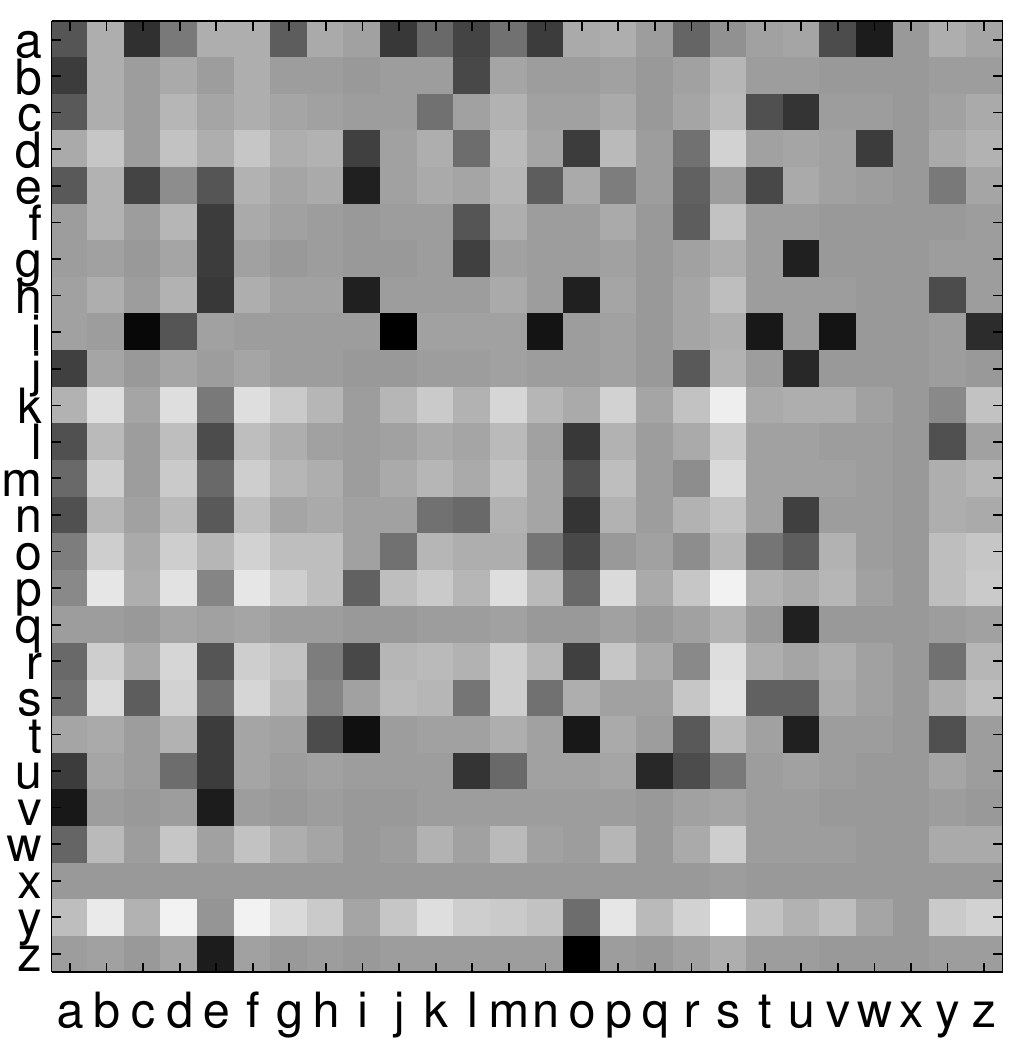} &
    \includegraphics[height=0.17\linewidth]{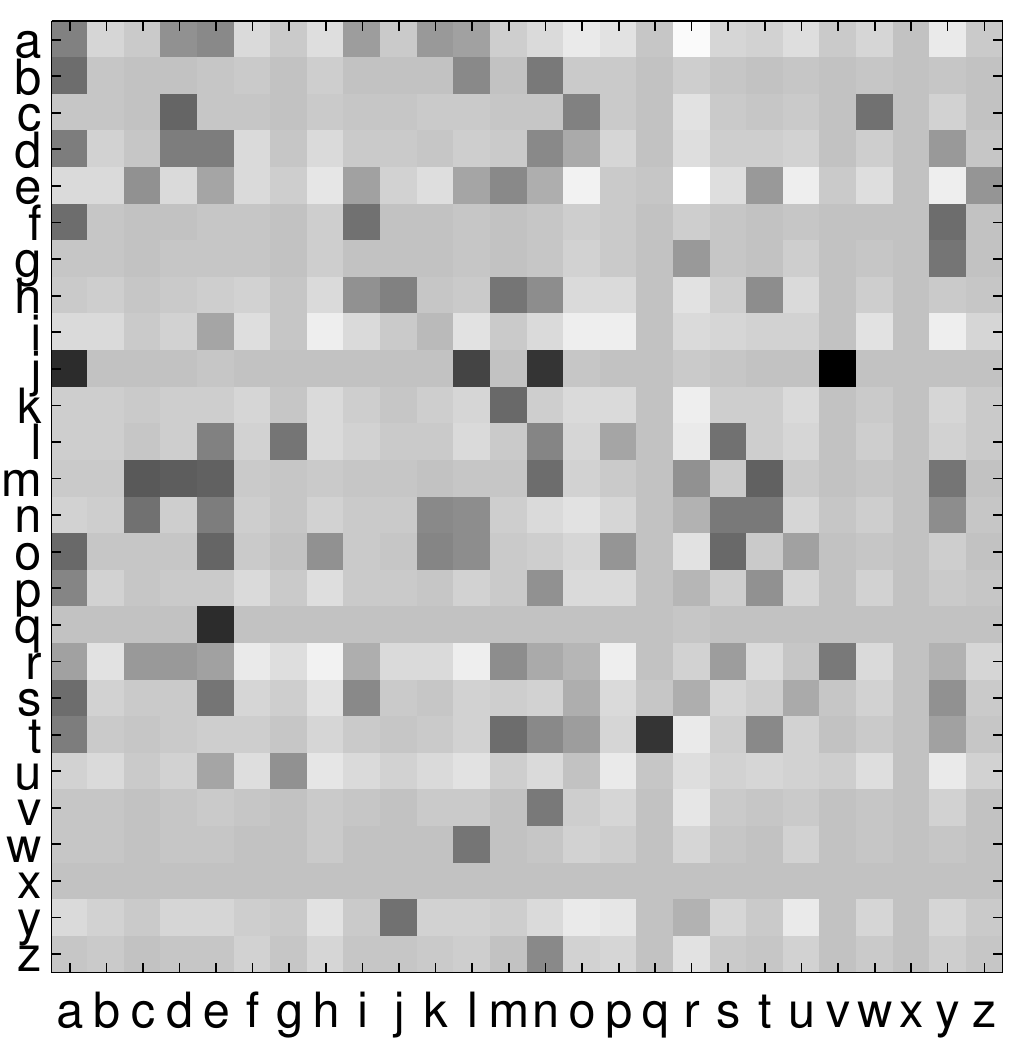} &
    \includegraphics[height=0.17\linewidth]{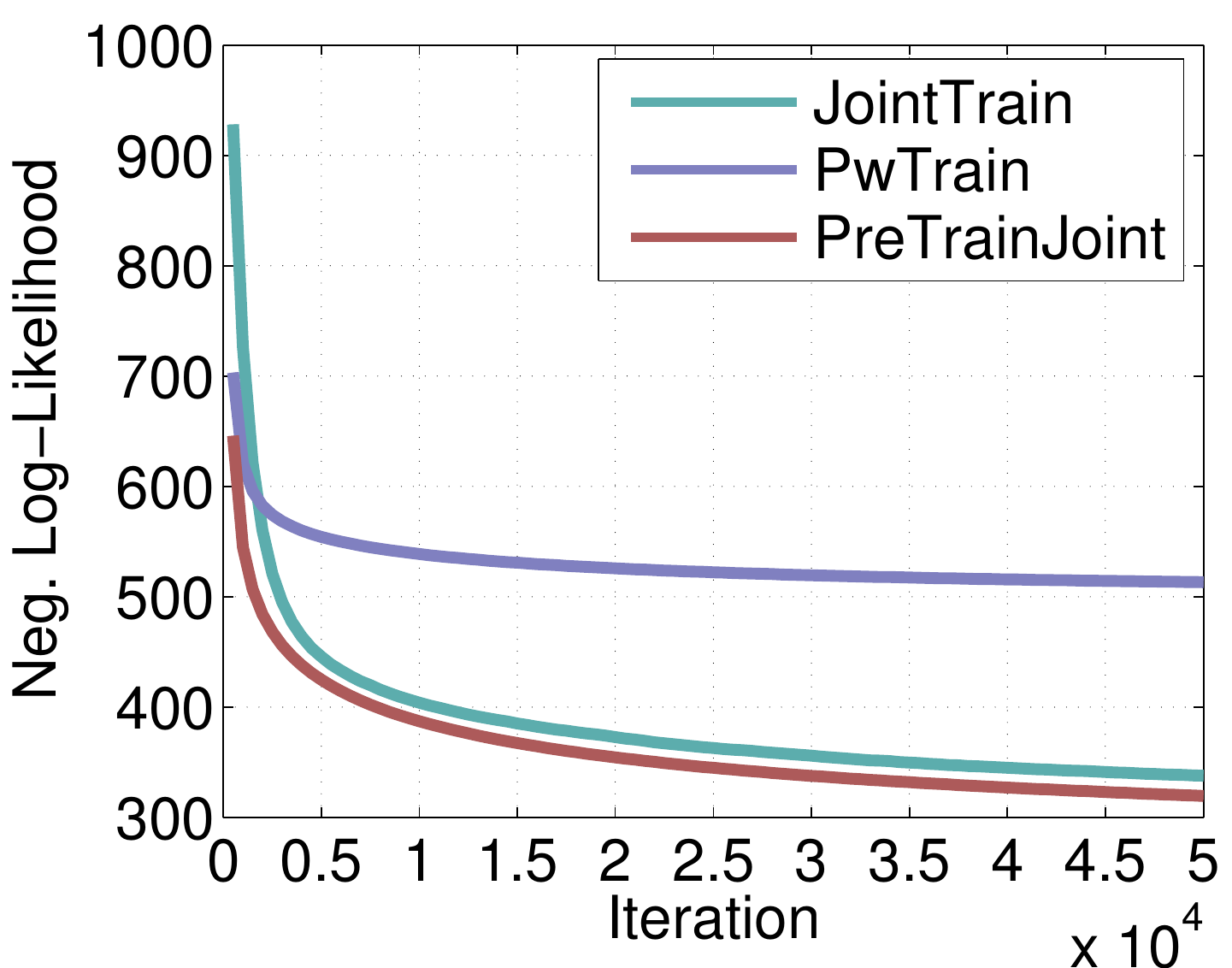} \\
    Unary weights & distance-1 edges & distance-2 edges & Neg. Log-Likelihood
  \end{tabular}
  \vspace{-0.0cm}
  \caption{ (left) Subset of the learned unary weights. Pairwise weights (middle two panels), the darker, the larger the weight. (right) Negative log-likelihood for different learning approaches. }
  \label{fig:word_filters}
	\vspace{-0.0cm}
\end{figure*}

\subsection{Word Recognition: Word50}

Our first task consists of word recognition from noisy images.  
Towards this goal, we created a challenging dataset by randomly selecting 50 words, each consisting of five characters. 
We then generated writing variations of each word as follows: we took the lower case characters from the Chars74K dataset~\citep{deCampos09}, and inserted them in 
 random background image patches (similar to \citet{larochelle2007empirical}) by alpha matting, \ie, characters have transparency.   
 To increase the difficulty, we perturbed each character image of size $28\times 28$ by scaling, rotation and translation. As shown  in  \figref{fig:WordDataSet} the task is very challenging, some characters are fairly difficult to recognize even for humans. We denote the resulting dataset `{\bf Word50}.'  The training, validation and test sets have $10,000$, $2,000$ and $2,000$ variations of words respectively. 

We experimented with graphical models composed of unary and pairwise regions defined over five random variables, one per character. 
We encode unary potentials $f_{r}(x, y_i; w_u)$ using multi-layer perceptrons (MLPs) with rectified linear units (ReLU). 
Unless otherwise stated, we define all pairwise interactions via 
\begin{align}
  \label{eqn:lin_pw}
  f_{r}(x, {y}_i,{y}_j;w_p) = \sum_{mn} W_{mn} \cdot \delta({y}_i=m,{y}_j=n),
\end{align}
where $r = \{i,j\}$, $w_p=\{W\}$, $W_{mn}$ is the element of matrix $W$, and $\delta$ refers to the indicator function.




\begin{figure}[t]
  \centering
  \addtolength{\tabcolsep}{-5pt}
  \begin{tabular} {c c}
    \includegraphics[height=0.25\linewidth]{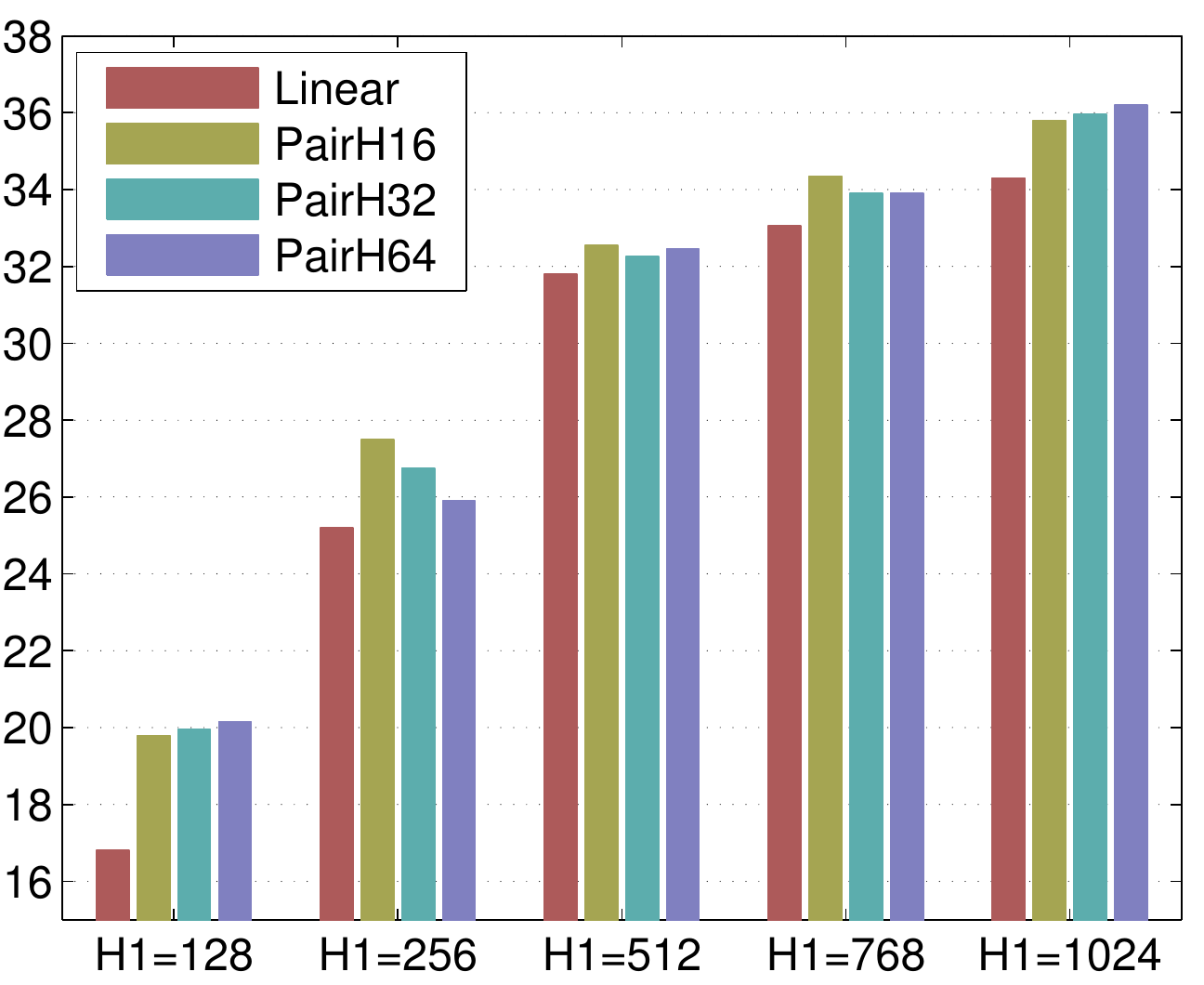} &
    \includegraphics[height=0.25\linewidth]{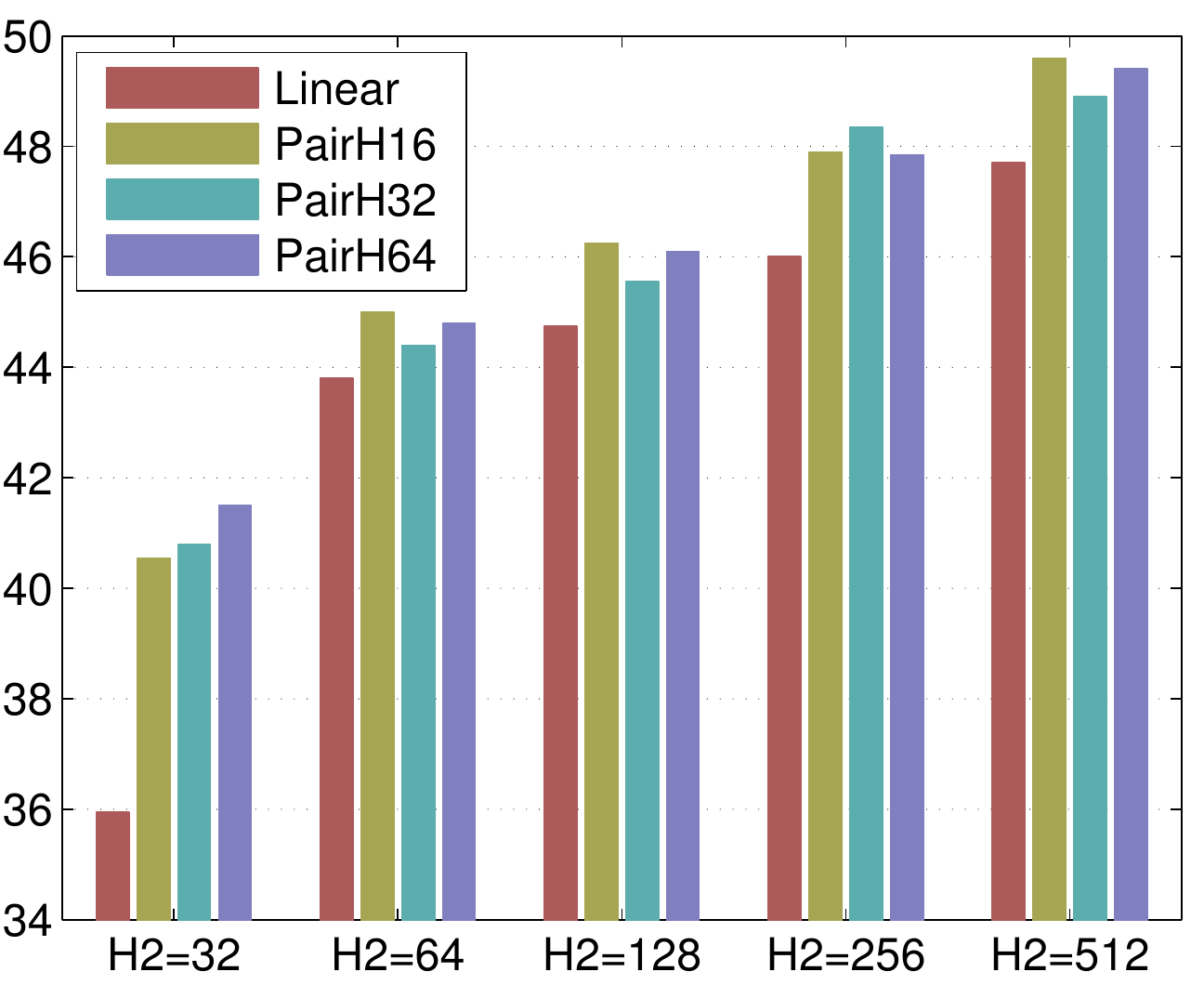} \\
    One-Layer MLP Chain & Two-Layer MLP Chain
  \end{tabular}
  \vspace{-0.0cm}
  \caption{Learning non-linear pairwise functions: Word recognition as a function of the number of hidden units for the unary potential. Colors represent different number of hidden units for the pairwise potentials. The y-axis shows the word accuracy of using Linear function, or 16 (PairH16), 32 (PairH32), and 64 (PairH64) hidden units for the pairwise function.
  }
  \label{fig:ChainPwNonLin}
	\vspace{-0.0cm}
\end{figure}

For all experiments, we share all unary weights across the nodes of the graphical model as well as all pairwise weights for all edges. 
Note that due to the use of ReLU units, the negative log-likelihood is non-smooth, non-linear and  non-convex \wrt $w$. 
Because of the non-smoothness of $F$, we utilize   momentum based sub-gradient descent methods to estimate the weights. In particular, we use a mini-batch size of $100$, a  step size of  $0.01$ and a momentum of $0.95$. If the unary potential is pre-trained, the initial step size is reduced to $0.001$. All the unary classifiers are trained with $100,000$ iterations over mini-batches. 
For all experiments, the validation set is only used to decrease the step size, \ie, if the accuracy on the validation set decreases, we reduce the step size by $0.5$. 
We use $\epsilon = 1$, set $c_r = 1$ for all regions $r$, and perform 10 message passing iterations to compute the marginal beliefs $b_{(x,y),r}$ at step 2 in \figref{fig:NonLinStructPredAlgo} when dealing with loopy models. 


We experiment with two graphical models, Markov models of first  (\ie,  there are links only between  $y_i$ and $y_{i+1}$) and second order (\ie, there are links between $y_i$ and  $y_{i+1}$,  $y_{i+2}$) as well as  two types of unary potentials 
with varying degree of structure. 
We report two metrics, the average character  and  word accuracy, which correspond to  Hamming loss and zero-one loss respectively. 
\tabref{tab:ExpWordsDist1} depicts the results for the different models, learning strategies and number of hidden units. We observe the following trends.
 
 

\begin{figure*}[t]
\vspace{-0.0cm}
\centering
\begin{tabular}{ccccc}
\includegraphics[width=2.2cm]{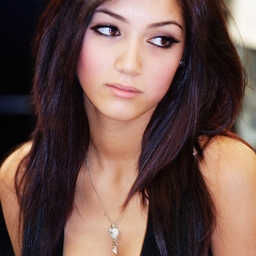}&\hspace{-0.15cm}\includegraphics[width=2.2cm]{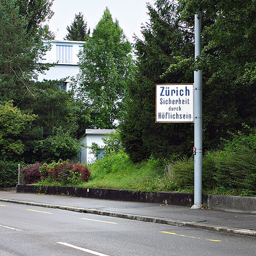}&\hspace{-0.15cm}\includegraphics[width=2.2cm]{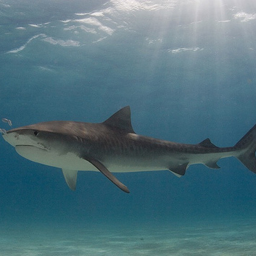}&\hspace{-0.15cm}\includegraphics[width=2.2cm]{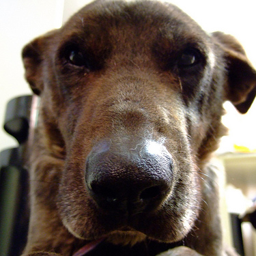}&\hspace{-0.15cm}\includegraphics[width=2.2cm]{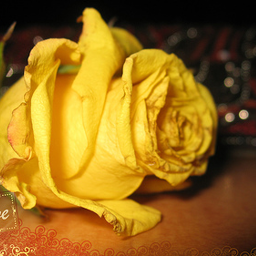}\\
{\small female/indoor/portrait}&\hspace{-0.15cm}{\small sky/plant life/tree}&\hspace{-0.15cm}{\small water/animals/sea}&\hspace{-0.15cm}{\small animals/dog/indoor}&\hspace{-0.15cm}{\small indoor/flower/plant life}\\
{\small female/indoor/portrait}&\hspace{-0.15cm}{\small sky/plant life/tree}&\hspace{-0.15cm}{\small water/animals/sky}&\hspace{-0.15cm}{\small animals/dog}&\hspace{-0.15cm}{\small $\emptyset$}\\
\end{tabular}
\vspace{-0.0cm}
\caption{Flickr test set images and some assigned tags as well as our predictions (bottom row).}
\label{fig:FlickrImages}
\vspace{-0.0cm}
\end{figure*}

\noindent{\bf Joint training helps:}
Joint training with pre-trained unary classifiers (PreTrainJoint) outperforms all the other approaches in almost all cases. The piecewise training method (PwTrain), unable to adapt the non-linearities  while learning pairwise weights, often leads to performance worse than joint training.



\noindent{\bf Structure helps:}
Adding structure to the model is  key to capture  complex dependencies. As shown in \tabref{tab:ExpWordsDist1},  more structured models (\ie, second order Markov model) consistently   improves performance. 

\paragraph{Deep helps:}
We  tested our models using one layer and  two-layer perceptrons with both short-range and long-range connections in the MRF. For the two-layer MLP, the number of hidden units in the first layer is fixed to $H_1 = 512$, and we varied the number of hidden units $H_2$ in the second layer. 
 As shown in \tabref{tab:ExpWordsDist1} we observe that the deeper and the more structured the model, the better the performance we achieve.
As expected, performance also grows with the number of hidden units. 

\noindent{\bf Efficiency:}  Using GPUs, it takes on average 0.064s per iteration for the 1st order Markov model and 0.104s for the 2nd order Markov model. The time employed for training the one layer \vs the multi-layer models is approximately the same. Note that our approach is very efficient, as this is the time per iteration to train 831,166 weights.

\noindent{\bf Learned parameters:}
As shown in the left column of \figref{fig:word_filters}, the learned unary weights resemble character strokes. 
The middle two panels show the learned pairwise weights for distance-1 edges (\ie, edges with only neighboring connections) and distance-2 edges (\ie, edges connecting every other variable). For example, it shows that `q' is likely to be followed by 	`u,' and `e' is likely to be distance-2 away from `q' in this dataset.
On the right-most panel, we also show the negative log-likelihood as a function of  the number of joint training iterations. 
PreTrainJoint can achieve the lowest cost value, while PwTrain has the highest value.

\noindent{\bf Non-linear pairwise functions:}
To further demonstrate the generality of our approach, we replaced the linear pairwise function in \equref{eqn:lin_pw} by a one-layer MLP, while keeping the other settings  identical. For this experiment we utilize a 1st order Markov model. 
As shown in \figref{fig:ChainPwNonLin}, our model attains best performance when using a non-linear pairwise function. We found 16 to 64 hidden units for the non-linear pairwise function to be sufficient for modeling the  bi-gram combinations in this dataset. In this case the largest model has  974,846 weights and training takes on average 0.068s per iteration. 

\begin{figure*}[t]
\vspace{-0.0cm}
\centering
\begin{tabular}{cc}
\hspace{-0.5cm}
\includegraphics[height=5.5cm]{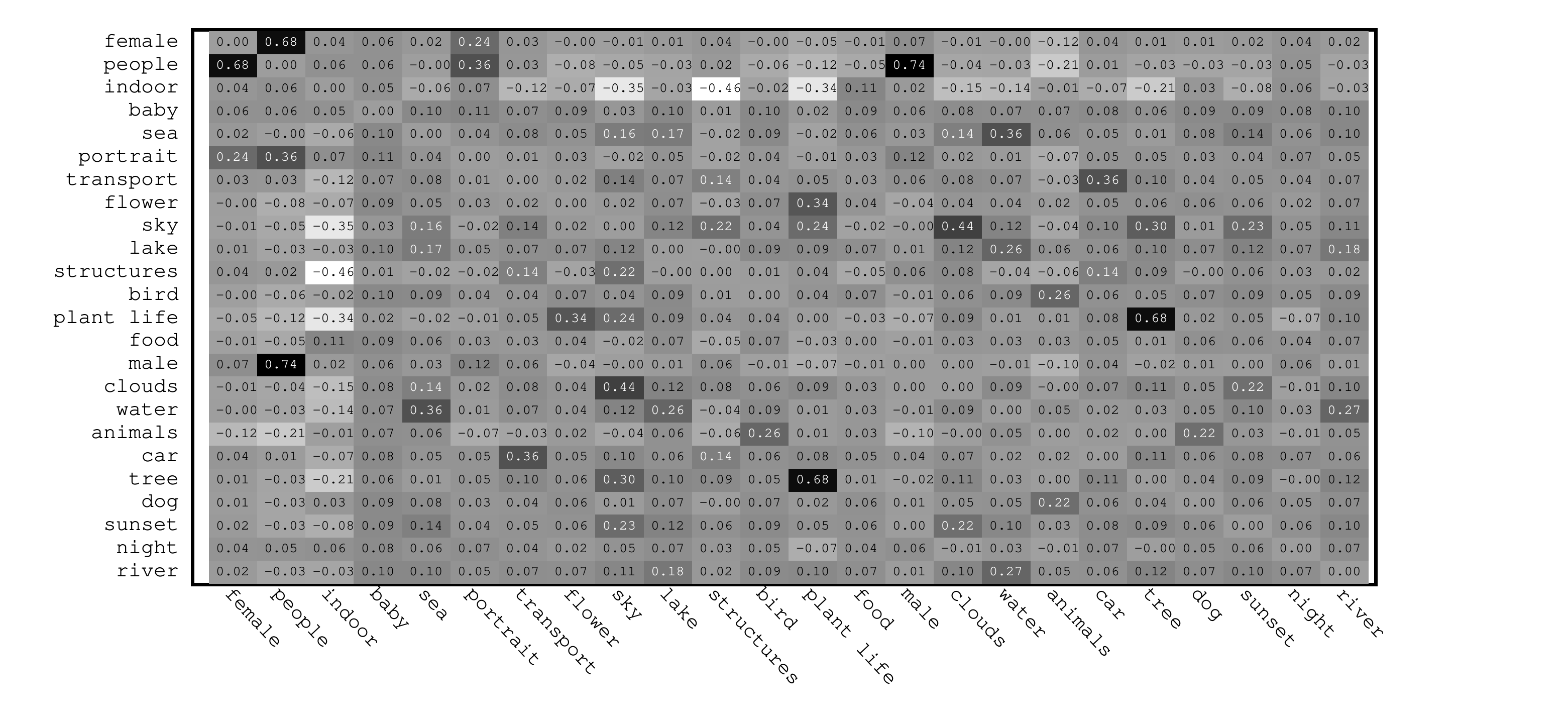}\vspace{-0.3cm}
&
\hspace{-1.5cm}\raisebox{2.9cm}{
\begin{tabular}{c}
\includegraphics[height=2.2cm]{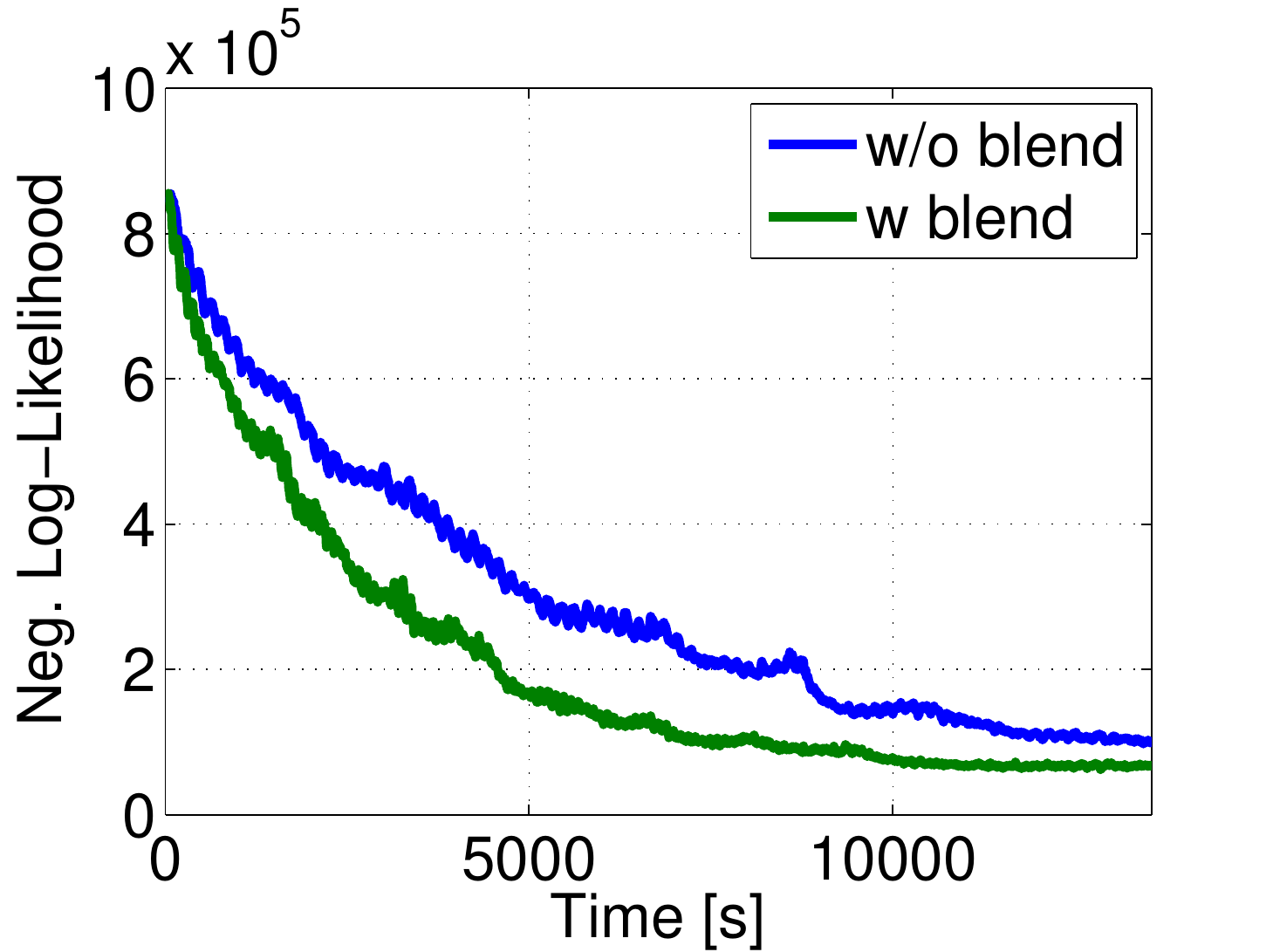}\\
\includegraphics[height=2.2cm]{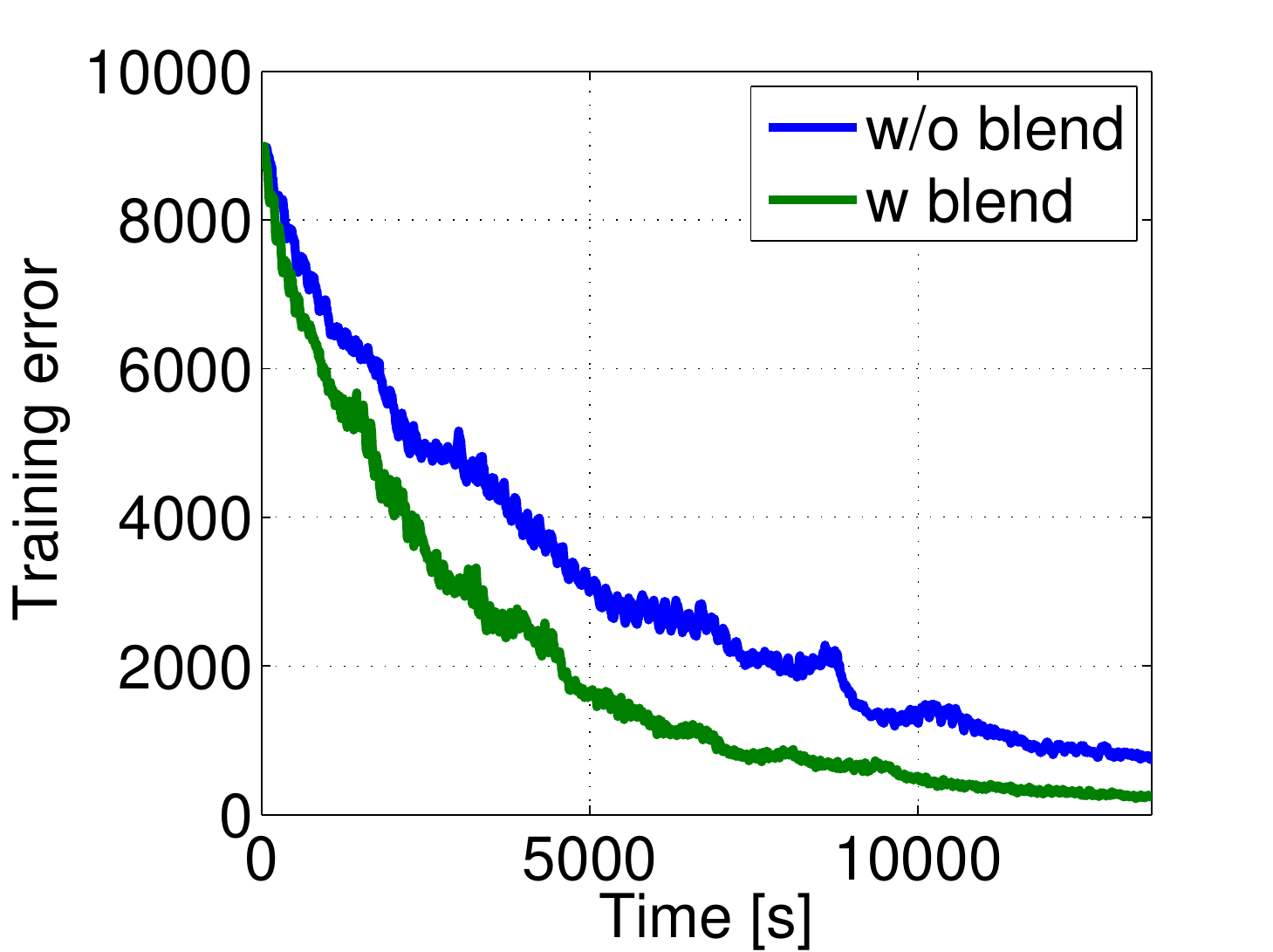}
\end{tabular}}\\
(a)&
\hspace{-1.5cm}(b)
\end{tabular}
\vspace{-0.0cm}
\caption{(a) Correlation matrix (\ie, pairwise potentials) learned on the Flickr dataset. (b) Blending learning and inference speeds-up training significantly.}
\label{fig:FlickrClassCorrelation}
\vspace{-0.0cm}
\end{figure*}

\subsection{Image Classification: Flickr}

We next evaluate the importance of blending learning and inference.  Towards this goal, we make use of the Flickr dataset, which consists of $10,000$ training and $10,000$ test images from  Flickr. The task is to predict which of 38 possible tags should be assigned to each image. \figref{fig:FlickrImages} shows some examples.  
The graphical model has 38 binary random variables, each denoting the presence/absence of a particular tag. 
We define the  non-linear unaries $f_r(x,y_i; w_u)$ using  the 8-layer deep-net architecture from \citet{KrizhevskyNIPS2013} followed by a $76$-dimensional top layer. 
Hence the function is composed out of two subsequent stacks of convolution, rectified linear (ReLU), pooling and local response normalization units. Those are followed by three convolution--ReLU function pairs. Afterwards pooling is applied before two fully-connected--ReLU--dropout combinations are employed to yield the input into a fully connected layer which finally computes the unary potentials.
We employ pairwise potentials similar to \equref{eqn:lin_pw} which now fully model the correlations between any pair of output variables. 
This amounts to a total of $57,182,408$  parameters arising from the convolutional units, fully connected units and corresponding biases as well as the pairwise weights.

We  use a momentum based sub-gradient method for training with a mini-batch size of $300$, a step size of $0.0001$,  a momentum of $0.95$ and set $\epsilon = 1$ and $c_r = 1$ $\forall r$. 
We initialized the deep-net parameters using a model pre-trained on ImageNet \citep{deng2009imagenet}.
Our error metric is the classification error (\ie, Hamming loss). 

\noindent{\bf Joint training helps:} 
The mean error for unary only potentials (`Unary only'), piecewise training (`PwTrain') and joint pretraining (`PreTrainJoint') is 9.36\%, 7.70\% and 7.25\% respectively. 
Similar to the Word50 dataset we observe that joint training is beneficial. We provide examples  for perfect (two left-most images), roughly accurate and failing predictions (right image) in \figref{fig:FlickrImages}. 

\noindent{\bf Learned pairwise weights:} In \figref{fig:FlickrClassCorrelation}(a) we illustrate the learned correlations for a subset of the 38 classes. We observe that the class `people' correlates highly with `female,' `male,' and `portrait.' The `indoor' tag  does not co-occur with `sky,' `structures,' `plant life' and `tree.'  `Sea' appears typically with  `water,' `clouds,' `lake' and `sky.'

\noindent{\bf Efficiency of Blending:}
To illustrate that blending is indeed beneficial we compare the negative log-likelihood and the training error as a function of run-time in \figref{fig:FlickrClassCorrelation}(b).  
The standard approach is limited to 20 iterations of message passing    to avoid time-consuming, repeated computation of a stopping criterion involving both the approximated log-partition function and its dual.
As show in \figref{fig:FlickrClassCorrelation}(b) blending learning and inference speeds up parameter estimation significantly. 
For larger graphical models, we expect the differences to be even more significant.


\section{Discussion \& Conclusion}

{\bf{Joint training of neural networks and graphical models:}} Neural Networks have been incorporated as unary potentials in graphical models. One of the earliest works by \citet{bridle1990training} jointly optimizes a system consisting of multilayer perceptrons and hidden Markov models for speech recognition. For document processing systems, \citet{bottou1997global} propose Graph Transformer Networks to jointly optimize sub-tasks, such as word segmentation and character recognition. Several works \citep{collobert2011natural, PengNIPS2009, MaBioinformatics2012, do2010neural, prabhavalkar2010backpropagation, morris2008conditional} have extended the linear unary potential in MRFs to incorporate non-linearities. 
However, they assume that exact inference can be performed either via a forward-backward pass within the graphical model or dynamic programming. In contrast, in this paper we present learning algorithms for  general graphical models, where inference is hard. 
Moreover, all the previous works (except \citet{do2010neural}) do not consider max-margin loss during training which is incorporated into our framework by choosing $\epsilon = 0$. 
More recently, \citet{LiICML2014} use a hinge loss to learn the unary term defined as a neural net, but keep the pairwise potentials fixed (\ie, no joint training). \citet{domke2013structured} considers non-linear structured prediction and decomposes the learning problem into a subset of logistic regressors, which  require the parameter updates to be run till convergence before updating the messages. \citet{tompson2014joint} also jointly train convolutional neural networks and a graphical model for pose estimation. However, the MRF inference procedure is approximated by their Spatial-Model  which ignores the partition function.


{\bf{Blending learning and inference:}} In this paper we defined learning to be a $\min$-$\max$ task. The blending strategy, which was previously employed for learning log-linear models by  \citep{Meshi2010,Hazan2010b}, amounts to converting the maximization task into a minimization problem using its dual. Subsequently we make use of block-coordinate descent strategies to obtain a more efficient algorithm. Importantly any order of block-updates is possible. It remains an open problem to find the optimal tradeoff. 



We have proposed an efficient algorithm to learn deep models enriched to capture the dependencies between the output variables.  
Our experiments on word prediction from noisy images and multi-class image classification showed that  the deeper and the more structured the model, the better the performance we achieve. Furthermore, joint learning of all weights outperforms all other strategies. 
In the future we plan to learn deeper models in  applications such as holistic semantic scene understanding. We will also extend our approach to deal with hidden variables.

\subsubsection*{Acknowledgments}
We thank NVIDIA Corporation for the donation of GPUs used in this research. This work was  partially funded by ONR-N00014-14-1-0232.

{\small
\bibliography{egbib}
\bibliographystyle{iclr2015}
}

\end{document}


\twocolumn[

\aistatstitle{Supplementary material for Learning Deep Structured Models}

\aistatsauthor{ Anonymous Author 1 \And Anonymous Author 2 \And Anonymous 
Author 3 }
\aistatsaddress{ Unknown Institution 1 \And Unknown Institution 2 \And Unknown Institution 3 
} ]

In this supplementary material, we show test prediction results on the Word50 dataset. In particular, we show  prediction results for single-layer MLP with $H_1=1024$, and two-layer MLP with $H_1=512$  and $H_2=512$. The goal is to demonstrate the improvements as a function of the amount of structure captured in the graphical model. We denote the single-layer MLP with 1st order Markov model, single-layer MLP with 2nd order Markov model, two-layer MLP with 1st order Markov model, and two-layer MLP with 2nd order Markov model as `{\bf One-Chain}', `{\bf  One-Long}', `{\bf Two-Chain}', and `{\bf Two-Long}' respectively. For comparison, we also provide the ground-truth and results when using only single-layer MLP and two-layer MLP as unary functions, denoted as `{\bf GT}', `{\bf One-Unary}', and `{\bf Two-Unary}'.  We show success modes from \figref{fig:SuccPredictionResults_1} to \figref{fig:SuccPredictionResults_5}, and failure modes from \figref{fig:FailPredictionResults_1} to \figref{fig:FailPredictionResults_5}. As shown in the figures, the more structure the graphical model captures and the deeper the model is, the better the performance attained. For the failure modes, some words are very hard to distinguish, even for humans.

\clearpage

\begin{figure*}[h] 
\vspace{-0.2cm}
\hspace{-1.4cm}

\vspace{-0.3cm}
\caption{Successful prediction results from the Word50 dataset.
} 
\label{fig:SuccPredictionResults_1}
\end{figure*}\begin{figure*}[h] 
\vspace{-0.2cm}
\hspace{-1.4cm}
%
\vspace{-0.3cm}
\caption{Successful prediction results from the Word50 dataset.
} 
\label{fig:SuccPredictionResults_2}
\end{figure*}\begin{figure*}[h] 
\vspace{-0.2cm}
\hspace{-1.4cm}
%
\vspace{-0.3cm}
\caption{Successful prediction results from the Word50 dataset.
} 
\label{fig:SuccPredictionResults_3}
\end{figure*}\begin{figure*}[h] 
\vspace{-0.2cm}
\hspace{-1.4cm}
%
\vspace{-0.3cm}
\caption{Successful prediction results from the Word50 dataset.
} 
\label{fig:SuccPredictionResults_5}
\end{figure*}

\begin{figure*}[h] 
\vspace{-0.2cm}
\hspace{-1.4cm}
%
\vspace{-0.3cm}
\caption{Failure prediction results from the Word50 dataset.
} 
\label{fig:FailPredictionResults_1}
\end{figure*}\begin{figure*}[h] 
\vspace{-0.2cm}
\hspace{-1.4cm}
%
\vspace{-0.3cm}
\caption{Failure prediction results from the Word50 dataset.
} 
\label{fig:FailPredictionResults_2}
\end{figure*}\begin{figure*}[h] 
\vspace{-0.2cm}
\hspace{-1.4cm}
%
\vspace{-0.3cm}
\caption{Failure prediction results from the Word50 dataset.
} 
\label{fig:FailPredictionResults_3}
\end{figure*}\begin{figure*}[h] 
\vspace{-0.2cm}
\hspace{-1.4cm}
%
} & shete & snere & snere & snere & snete & snete & snete\\ 
\hline
\end{tabular}
\vspace{-0.3cm}
\caption{Failure prediction results from the Word50 dataset.
} 
\label{fig:FailPredictionResults_5}
\end{figure*}